\documentclass[letterpaper, 10 pt, conference]{ieeeconf}  % Comment this line out if you need a4paper

\IEEEoverridecommandlockouts                              % This command is only needed if 
\usepackage{fancyhdr}
\fancypagestyle{arxivhdr}
{
   \fancyhf{}
   \setlength{\headheight}{15pt} 
\fancyfoot[C]{This paper has been accepted for publication at the 2024 IEEE/RSJ International Conference on\\Intelligent Robots and Systems (IROS)}
\fancyhead[C]{\footnotesize Please cite this paper as:\\
D. Fusaro, S. Mosco, E. Menegatti and A. Pretto, "Exploiting Local Features and Range Images for Small Data Real-Time Point Cloud Semantic Segmentation," 2024 IEEE/RSJ International Conference on Intelligent Robots and Systems (IROS), 2024}
}

\title{\LARGE \bf
Exploiting Local Features and Range Images for Small Data \\ Real-Time Point Cloud Semantic Segmentation
}

\usepackage{url}
\usepackage{cite}
\usepackage{amsmath,amssymb,amsfonts}
\usepackage{algorithmic}
\usepackage{graphicx}
 \usepackage{array,multirow,float}
\usepackage{textcomp}
\usepackage{xcolor}
\usepackage{pifont}
\usepackage{balance}

\newcommand\new[1]{#1}
\newcommand{\cmark}{\ding{51}}%
\newcommand{\xmark}{\ding{55}}%

\newcommand{\mycomment}[1]{}

\usepackage{rotating} % for rotating text

\newcommand{\rb}[1]{% Define a custom command for overlapping rotated text
  \rotatebox[origin=l]{70}{#1}\hspace{-0.5mm} % Rotate with left origin and adjust spacing
}

\author{Daniel Fusaro, Simone Mosco, Emanuele Menegatti and Alberto Pretto % <-this % stops a space
\thanks{\{fusarodani, moscosimon, emg, alberto.pretto\}@dei.unipd.it
}
\thanks{Department of Information Engineering, University of Padova, Italy. %
}
}

\begin{document}

\maketitle
% Arxiv header
\thispagestyle{empty}
\pagestyle{empty}
\thispagestyle{arxivhdr}% Title page has fancy

%%%%%%%%%%%%%%%%%%%%%%%%%%%%%%%%%%%%%%%%%%%%%%%%%%%%%%%%%%%%%%%%%%%%%%%%%%%%%%%%
\begin{abstract}

Semantic segmentation of point clouds is an essential task for understanding the environment in autonomous driving and robotics. Recent range-based works achieve real-time efficiency, while point- and voxel-based methods produce better results but are affected by high computational complexity. Moreover, highly complex deep learning models are often not suited to efficiently learn from small datasets. Their generalization capabilities can easily be driven by the abundance of data rather than the architecture design. 
In this paper, we harness the information from the three-dimensional representation to proficiently capture local features, while introducing the range image representation to incorporate additional information and facilitate fast computation.
% We make use of a GPU-based KDTree for rapid building and querying and enhance projection with straightforward operations.
\new{A GPU-based KDTree allows} for rapid building, querying, and enhancing projection with straightforward operations.
Extensive experiments on SemanticKITTI and nuScenes datasets demonstrate the benefits of our modification in a ``small data'' setup, in which only one sequence of the dataset is used to train the models, but also in the conventional setup, where all sequences except one are used for training.
% A reduced version of our model not only demonstrates strong competitiveness against full-scale state-of-the-art models but also operates in real-time, rendering it a viable choice for real-world case applications.
\new{We show that a reduced version of our model not only demonstrates strong competitiveness against full-scale state-of-the-art models but also operates in real-time, making it a viable choice for real-world case applications.}
The code of our method is available at \url{https://github.com/Bender97/WaffleAndRange}. 
\end{abstract}

%%%%%%%%%%%%%%%%%%%%%%%%%%%%%%%%%%%%%%%%%%%%%%%%%%%%%%%%%%%%%%%%%%%%%%%%%%%%%%%%
\section{INTRODUCTION}
Point cloud semantic segmentation 
is a challenging task that has gained considerable interest in recent years, mainly due to its application in relevant fields such as autonomous driving, robotics, and environmental perception \cite{guo2020deep}. 
In particular, accurate semantic segmentation of the surrounding environment is essential for decision-making even in real-time applications such as autonomous vehicle navigation and intelligent robot navigation.
However, achieving a balance between accuracy and computational efficiency remains a considerable challenge. An even greater challenge lies in achieving accurate semantic segmentation when the available dataset is small. In such cases, the learning model must demonstrate exceptional generalization capabilities to accurately segment unseen data. Generally, highly complex state-of-the-art models tend to perform comparably well or even worse than low-complexity neural networks  \cite{9412492}.
%Accurate semantic segmentation understanding of the surrounding environment is essential for decision-making in applications such as autonomous vehicles and intelligent robotics. However, achieving a balance between accuracy and computational efficiency, particularly in real-time scenarios, remains a considerable challenge.

Recent approaches have made significant advancements in the field by leveraging various LiDAR point cloud representations, including point-based \cite{thomas2019kpconv, hu2020randla, yan2020pointasnl}, voxel-based \cite{zhu2021cylindrical}, pseudo-image \cite{milioto2019rangenet++, ando2023rangevit, kong2023rethinking}, and hybrid representation \cite{xu2021rpvnet, hou2022point, yan20222dpass}. 
Notably, hybrid representations generally outperform other classes of methods by taking advantage of the strengths of each representation.
%Notably, hybrid representations stand out for their capacity to enhance accuracy by capitalizing on the strengths of each representation.
However, while the richness of information benefits performance, it introduces drawbacks in terms of processing time, making them unsuitable for real-world scenarios. 

Range image representations allow the use of established 2D convolutional neural networks (CNNs), which require less processing time but often produce less accurate results due to information loss caused by the projection of 3D points onto a 2D plane. In contrast, point and voxel-based methods benefit from the original 3D representation, enabling the capture of local features and point distribution. However, these methods require significantly longer computation times. 
In this work, we rely on both the aforementioned representations by combining the strengths of two recent approaches, WaffleIron \cite{puy2023using} and RangeFormer \cite{kong2023rethinking}, to achieve real-time efficiency and enhance generalization capabilities. 
%In this work, drawing inspiration from WaffleIron \cite{puy2023using} and RangeFormer \cite{kong2023rethinking}, we combine the strengths of both approaches to achieve real-time efficiency.
Our network comprises three main components: a point cloud embedding module, a backbone composed of point cloud processing layers, and a segmentation head for generating final predictions. Specifically, we explore the utility of the point embedding module in WaffleIron, which encompasses richer information, operates at a three-dimensional level, and leverages local features, geometry, and shape characteristics. Simultaneously, we harness the range image in the backbone as a valuable representation, facilitating the use of dense 2D convolution and ensuring faster computational time.
\begin{figure}
\includegraphics[width=0.5\textwidth]{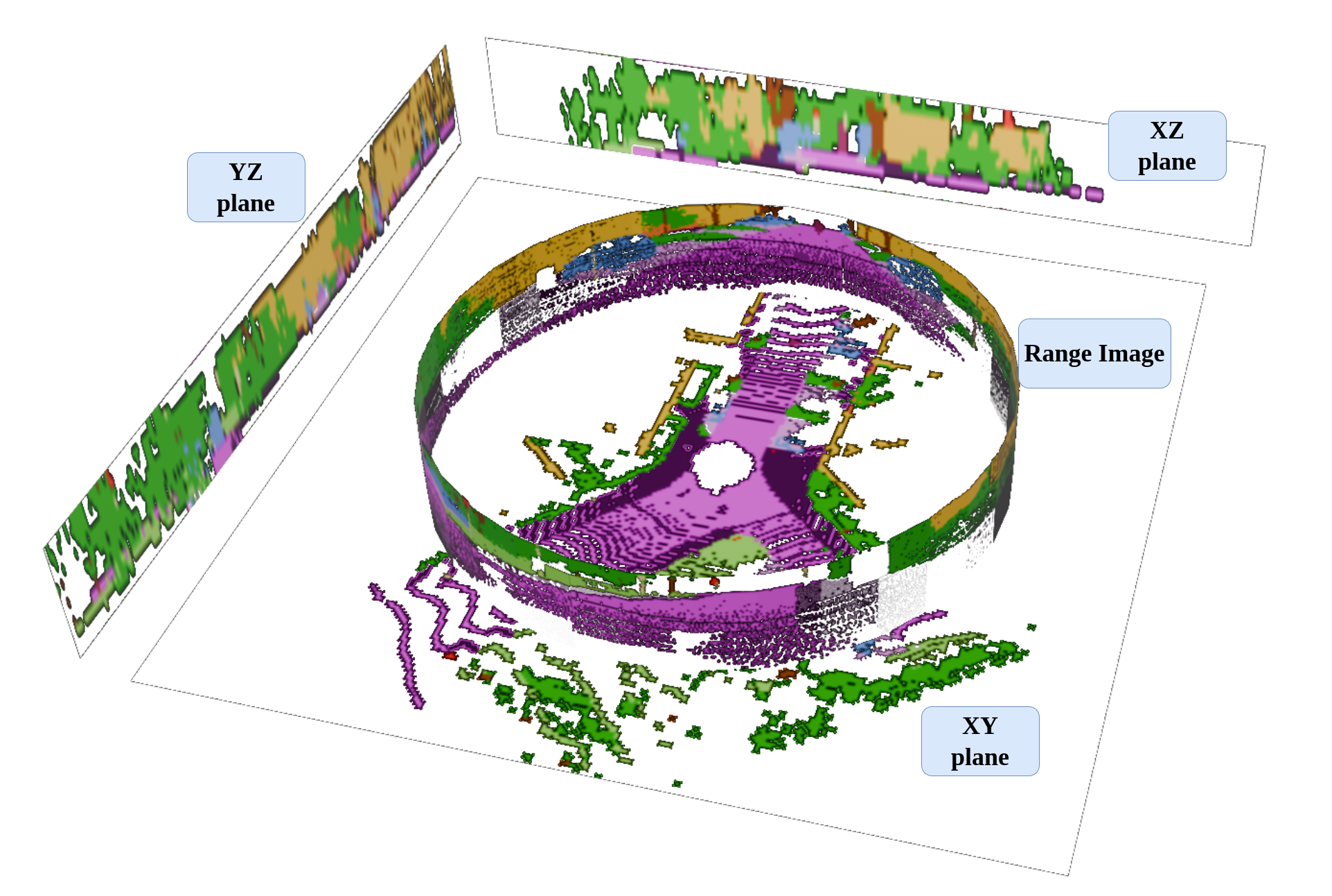}
\caption{The four 2D projections utilized by our system for semantically segmenting the 3D point cloud are as follows: XY, XZ, YZ, and range image projection.}
\label{fig:projections}
\end{figure}
In addition, we prioritize the optimization of inference time. We introduce a GPU-based KDTree for building and querying inside the point cloud embedding module, offering time savings compared to the CPU version. Furthermore, we enhance the flattening operations proposed in WaffleIron through straightforward element-wise multiplication and scattering operations. 
This ensures faster computation and lower memory consumption because we do not use large and very sparse matrices for this operation.

The key contributions of this work are as follows:
\begin{itemize}
%\item We propose a novel Deep Learning architecture for \new{enhancing the quality of semantic segmentation of } automotive point clouds \new{in little-data setups}, comprising standard MLPs and dense 2D convolutions, achieved by combining two state-of-the-art models: WaffleIron and Rangeformer, and optimizing this combination.
    \item We introduce a novel deep learning architecture integrating two state-of-the-art models, WaffleIron and Rangeformer, designed to improve the quality of semantic segmentation for automotive point clouds in scenarios with limited data availability. % This architecture combines standard Multi-Layer Perceptrons (MLPs) and dense 2D convolutions} %,integrating two state-of-the-art models: WaffleIron and Rangeformer.}
\item We demonstrate through extensive experiments that the extraction of local features within the embedding module, coupled with the integration of range image data, significantly enhances performance in scenarios characterized by limited data availability (hereinafter referred to as ``small data'' setup). Furthermore, our approach yields meaningful improvements even in contexts where abundant data is available, thus exhibiting its versatility across different data regimes.
%We optimize this combination to enhance the performance of semantic segmentation tasks in little-data setups.}
% \item We drastically reduce the model system runtime to just ~180 ms per point cloud without any inference optimizer, thanks to code logic optimization.
\item \new{We drastically reduce the full-model system runtime to just 180 ms per point cloud without any inference optimizer, thanks to code optimization and a GPU-based KDTree. We show that a reduced version of our model shows strong competitiveness with full-scale state-of-the-art models but also operates in real-time, with a runtime of 80 ms.}
\item We conduct an in-depth performance evaluation of our and other recent methods on two autonomous driving benchmarks: SemanticKITTI \cite{semantickittidataset} and nuScenes \cite{nuscenes}.
\item An open-source implementation of the proposed method is released for public usage.
\end{itemize}

\section{RELATED WORKS}

\begin{figure*}[ht]
\includegraphics[width=\textwidth]{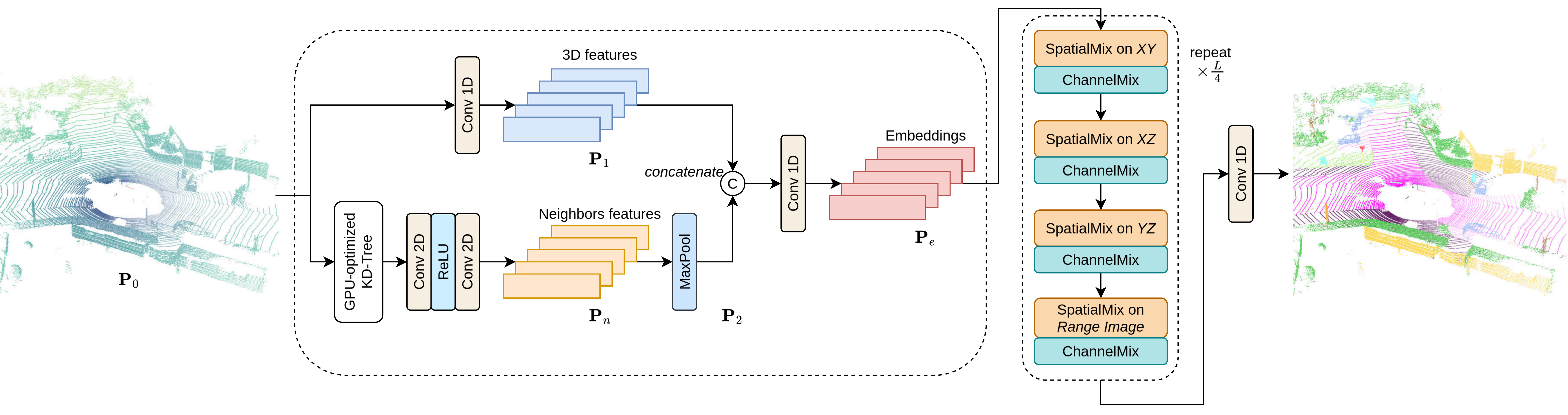}
% \end{figure*}
% \begin{figure*}[h]
\includegraphics[width=\textwidth]{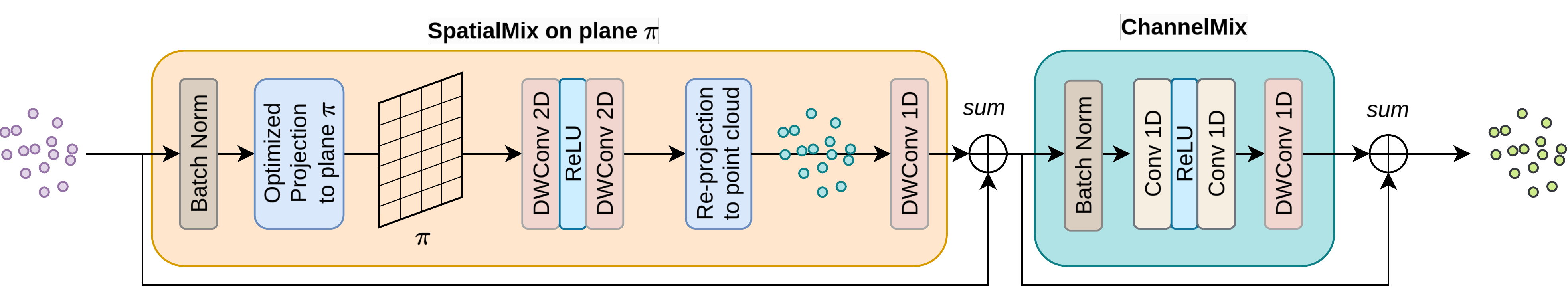}

\caption{(Above) An architectural overview of the proposed method featuring Point Cloud Embedding for point features computation, Point Cloud processing layers as the backbone with integrated Spatial and channel mix modules, and a Segmentation Head for generating final predictions. (Below) A detailed representation of the backbone, showcasing the spatial mix and channel mix modules. The spatial mix includes batch normalization, projection onto a 2D plane, 2D depth-wise convolution, re-projection onto 3D points, 1D depth-wise convolution, and a residual connection. Meanwhile, the channel mix employs batch normalization, 1D convolution, 1D depth-wise convolution, and a residual connection.}
\label{fig:network}

\end{figure*}

Point cloud semantic segmentation methods can be divided into four paradigms: \textit{point-based}, \textit{projection-based}, \textit{discretization-based}, and \textit{hybrid methods}.

\subsection{Point-based Methods}
Point-based methods directly work on the raw unordered point cloud. PointNet \cite{qi2017pointnet} is the pioneer in this field, followed by its improved version PointNet++ \cite{qi2017pointnet++}. They approximate a symmetric function using multi-layer perceptrons (MLPs) to obtain permutation invariance of the input points and learn per-point features. KPConv \cite{thomas2019kpconv}, DGCNN \cite{wang2019dynamic} and PointConv \cite{wu2019pointconv} apply convolution-like operations to exploit local geometric features. In RandLA-Net \cite{hu2020randla} the point cloud is first subsampled while an attention mechanism retains relevant information. PointASNL \cite{yan2020pointasnl} introduces an adaptive sampling module to deal with point clouds with noise. WaffleIron \cite{puy2023using} proposes a novel 3D backbone, relying on MLPs over 3D points and dense 2D convolution on projection planes. Recent works \cite{zhao2021point, wu2022point, wu2023point} rely on the self-attention mechanism introduced in transformers \cite{vaswani2017attention} and directly work on the 3D points. However, point-based methods are rarely used to process LiDAR point clouds since they are generally time-consuming.

\subsection{Projection-based Methods}
These methods usually rely on a 2D representation of the point cloud by projecting it onto a surface. SqueezeSeg \cite{wu2018squeezeseg} proposes an encoder-decoder network based on SqueezeNet \cite{iandola2016squeezenet} and conditional random fields to refine the predictions. SqueezeSegv2 \cite{wu2019squeezesegv2} and SqueezeSegv3 \cite{xu2020squeezesegv3} introduce respectively a context aggregation module and a spatially-adaptive convolution to improve the network. In \cite{caltagirone2017fast} a top-view image is extracted and fed to a fully connected network, while in \cite{zhang2020polarnet} bird's eye view projections are exploited. On the opposite side, SalsaNet \cite{aksoy2020salsanet} and its evolution SalsaNext \cite{cortinhal2020salsanext} claim that the projection type does not bring any advantage to the segmentation of their work. RangeNet++ \cite{milioto2019rangenet++} integrates DarkNet as the backbone to leverage range images and proposes an efficient k-NN post-processing technique. KPRNet \cite{kochanov2020kprnet} exploits KPConv as segmentation head while Lite-HDSeg \cite{razani2021lite} introduces three different modules and a boundary loss to improve the results. FIDNet \cite{zhao2021fidnet} and CENet \cite{cheng2022cenet} switch encoders to ResNet \cite{he2016deep} and substitute the decoders with simple interpolation. Recent works as RangeViT \cite{ando2023rangevit} exploit vision transformers \cite{dosovitskiy2020image} pre-trained on image data, while RangeFormer \cite{kong2023rethinking} achieves state-of-the-art performance introducing a pyramidal structure based on \cite{wang2021pyramid}. Projection-based approaches usually leverage accelerated computations by operating almost entirely within the 2D image space, thereby attaining real-time efficiency.

\subsection{Discretization-based Methods}
These methods first discretize the 3D point cloud into voxel representation and then leverage 3D convolution operators. In \cite{tchapmi2017segcloud} the input is voxelized but the label predictions are refined at a per-point level. MinkowskiNet \cite{choy20194d} uses sparse convolution \cite{graham20183d} to reduce the computational cost while (AF)$^2$-S3Net \cite{cheng2021af2s3net} aggregates multi-scale features with an attention mechanism. Cylinder3D \cite{zhu2021cylindrical} uses a cylindrical grid to partition the space and designs an asymmetrical network to predict labels. Recent studies \cite{lai2023spherical} consider the varying sparsity in LiDAR point clouds, introducing a self-attention mechanism with a radial window. Discretization-based methods leverage the geometric properties of the 3D space, yet they incur higher computational costs when dealing with outdoor LiDAR point clouds. 

\subsection{Hybrid Methods}
Recent trends combine multiple representations (points, projection images, and voxels) or RGB images from camera sensors to boost performance. In \cite{tang2020searching, zhang2020deep, park2023pcscnet} fine-grained features at point level are integrated with high-level voxel representations. RPVNet \cite{xu2021rpvnet} introduces a range-point-voxel fusion network that leverages information from the three distinct representations. PVKD \cite{hou2022point} improves the performance by introducing a point-voxel knowledge distillation module while 2DPASS \cite{yan20222dpass} uses RGB images to enrich semantic and structural information during the training process. Recent works as FRNet \cite{xu2023frnet} combine range image pixels and frustum LiDAR points to boost point-level predictions while UniSeg \cite{liu2023uniseg} enriches voxel, range, and point-based representations with color and texture information from RGB images. Hybrid approaches leverage the strengths of each representation and may achieve a balanced trade-off between accuracy and efficiency.

\section{METHOD} \label{sec:method}

Fig.~\ref{fig:network} provides a detailed overview of our method. Based on the recent state-of-the-art works WaffleIron \cite{puy2023using} and RangeFormer \cite{kong2023rethinking}, it leverages the best of the two paradigms to produce high-quality point cloud segmentation and reduce the runtime of the system. It consists of three main stages: the point cloud embedding (pre-processing), the feature sharing stage, and the final segmentation head.

\subsection{Point Cloud Embedding}
Both WaffleIron and RangeFormer consist of these three stages, at least from a high-level point of view, implementing them in different flavors. Based on theoretical arguments and empirical evidence (see Section \ref{sec:ablation}), we found that WaffleIron's embedding gives a much more informative representation than RangeFormer's. The latter pre-processing stage consists of a module of 3 multi-layer perceptrons (MLPs), called range embedding module (REM), that directly operates on the range image built from the input point cloud. It maps the range image from $\mathbb{R}^{B \times 6 \times H \times W}$, where $B$ is the batch size, $H$ and $W$ are the range image's height and width respectively, to a higher dimensional space, $\mathbb{R}^{B \times C \times H \times W}$. The three steps are $6$ to $64$, $64$ to $128$, and $128$ again to $128$. It also consists of batch norm and Gaussian error linear unit (GELU) activation in each of the three MLPs.
%and 6 represents the input features size of each pixel,

WaffleIron's pre-processing does not operate on the range image but directly on the input point cloud. Firstly, the input point cloud is voxelized and cropped to the sensor field of view, reducing sensor outliers and ignoring points that are too far away.
% At this point, we obtain a point cloud where each point is associated with a feature in $\mathbb{R}^{N \times 5}$, where $N$ is the number of points in the point cloud and 5 is the dimensionality of each point's feature, namely x, y, z, LiDAR intensity value, and distance from the sensor (calculated as the Euclidean distance).
% Let's denote this point cloud representation as $P_0$.
\new{At this point, we obtain a point cloud $\mathbf{P_0 }\in \mathbb{R}^{N \times 5}$, where $N$ is the number of points in the point cloud and 5 is the dimensionality of each point's feature, namely x, y, z, LiDAR intensity value, and distance from the sensor (calculated as the Euclidean distance).}
$\mathbf{P_0}$ is processed by a simple 1D Convolutional layer that maps the five features to $128$, resulting in $\mathbf{P_1}$.
Then, a $K$-Dimensional Tree (KDTree) is built over the point cloud $\mathbf{P_0}$. We use this data structure to efficiently retrieve the $K$-Nearest Neighbors (kNN) to each point and build a local feature. Similar to PointNet \cite{qi2017pointnet}, the local feature extraction captures information such as shape characteristics, local geometry, and point distribution patterns. This hierarchical processing enables the network to learn representations that are robust to variations in scale, orientation, permutation, and density.

The \new{tensor} point cloud of neighbors \new{$\mathbf{P_n} \in \mathbb{R}^{K \times N \times 5}$ is obtained by stacking, for each point in $\mathbf{P_0}$, the $K$ neighbors' features. As} depicted in Fig.~\ref{fig:network}, \new{$\mathbf{P_n}$} is processed by a series of batch norm, 2D convolutions, and rectified linear unit (ReLU) activation functions. Following this processing, a max pooling operation is applied over the $K$ neighbors to select only the maximum response. This operation preserves the most salient features while suppressing noise, enhancing the representation's robustness to noisy input, and contributing to the permutation and rotation invariance of the network. Let's denote the output of this processing as the point cloud \new{$\mathbf{P_2} \in \mathbb{R}^{N \times 128}$}, which has the same number of points and feature size as $\mathbf{P_1}$, thanks to the pooling operation.

Finally, the two point clouds $\mathbf{P_1}$ and $\mathbf{P_2}$ are concatenated and remapped to a feature size of $128$ using a 1D convolution. \new{Let this final point cloud be denoted as $\mathbf{P_e }\in \mathbb{R}^{N \times 128}$.}

\subsection{Leveraging 2D and 1D Convolutions}

The preprocessed point cloud \new{$\mathbf{P_e}$} is then iteratively parsed by a sequence of spatial mix and channel mix operations. These manipulate 2D projections of the 3D point cloud and simply employ 1D or 2D convolutions. 3D convolutions are intentionally avoided to reduce the complexity of the model and thus mitigate overfitting drawbacks. Simpler models are generally more stable and tend to better generalize to unseen datasets (see Section~\ref{sec:ablation} for an empirical evaluation).

In this setup, the 3D data processing stage is reduced to the simpler, faster, and more efficient processing of 2D images.
2D convolutions excel in image semantic segmentation by capturing spatial information and relationships within an image. Unlike fully connected layers, which treat pixels independently, 2D convolutions consider local patterns and structures through kernel convolution. This facilitates learning features such as edges, textures, and shapes vital for object segmentation. Moreover, shared weights of convolutional kernels ensure parameter efficiency and translation invariance, making them ideal for capturing hierarchical representations crucial in semantic segmentation tasks.

1D convolutions are not commonly used for image semantic segmentation tasks due to their inability to capture the intricate spatial relationships present in images. Images inherently possess complex two-dimensional structures with spatial arrangements of pixels, and 1D convolutions are more suited for tasks involving sequential data. Through kernel convolution across a signal, they can extract meaningful features and relationships, aiding tasks such as denoising.

The 2D projections (see Fig.~\ref{fig:projections}) of the point cloud comprise orthogonal projections onto the planes XY, XZ, and YZ, as implemented in the original work WaffleIron, augmented with the 2D range image as seen in RangeFormer. The former consists of simple 2D grids where 3D points are projected, and cells are vectors in $\mathbb{R}^{128}$ obtained by averaging the features of points within the same cell.
Traditionally, the latter, as seen in RangeFormer, represents the point cloud by encoding the distance of each point from the sensor frame. Typically, only the closest point in each cell is considered for range image construction, resulting in information loss. In our work, however, we adopt the concept of feature averaging, similar to the three orthogonal projections, by considering all points belonging to the same cell of the range image.

A sequence of $L$ independent but identical modules parses these 2D projections. Each layer processes a specific 2D projection using the following procedure:
\begin{itemize}
    \item Project the points onto the specified 2D plane (XY, XZ, YZ, or Range Image).
    \item Perform the spatial mix operation over the 2D projection.
    \item Conduct point cloud feature inflation, where 3D points within the same 2D cell inherit the features of that cell.
    \item Execute the channel mix operation over the inflated 3D point cloud points.
    
\end{itemize}

The spatial mix and channel mix modules are illustrated in Fig.~\ref{fig:network}.
The spatial mix module comprises two 2D depth-wise convolutions separated by a ReLU activation function, followed by a 1D grouped convolution.
Similarly, the channel mix module mirrors the spatial mix module, with the 2D convolutions replaced by a 1D convolution.
Both modules begin with an initial batch norm of the input and end with a skip connection consisting in the summing of the output of the module and the input. 

\subsection{Segmentation Head}
To fully exploit the local features collected in the preprocessing step, we merge the parsed point cloud with the kNN embedding by simply summing them up. This skip connection offers evident benefits, such as enhancing the contextualization of point segmentation and facilitating gradient updates to the embedding module during the backpropagation phase.

The final labels are inferred by applying a 1D convolution over the merged point cloud followed by a Softmax layer.

\section{EXPERIMENTS}

\begin{table*}[t!]
\begin{center}
\caption{Semantic segmentation performance on SemanticKITTI validation set (sequence 8) of methods trained only on Sequence 04. The \xmark~symbol indicates classes not present in the training set.}
\label{tab:val-little-data-skitti}
\begin{tabular}{c p{2.5mm} | p{2.5mm}p{2.5mm}p{2.5mm}p{2.5mm}p{2.5mm}p{2.5mm}p{2.5mm}p{2.5mm}p{2.5mm}p{2.5mm}p{2.5mm}p{2.5mm}p{2.5mm}p{2.5mm}p{2.5mm}p{2.5mm}p{2.5mm}p{2.5mm}c}
\hline 
\rb{method} & \multicolumn{1}{c}{\rb{\textbf{mIoU \%} }} &
\rb{car} & \rb{bicycle} & \rb{motorcycle} & \rb{truck} & \rb{other-vehicle} & 
\rb{person} & \rb{bicyclist} & \rb{motorcyclist} & \rb{road} & \rb{parking} & \rb{sidewalk} & \rb{other-ground} & \rb{building} &
\rb{fence} & \rb{vegetation}  & \rb{trunk} & \rb{terrain} & \rb{pole} & \rb{traffic-sign} \\ 
\hline
RangeFormer \cite{kong2023rethinking} & 15.6 & 47.9 & \xmark & \xmark & \xmark & 0.2 & 0.1 & \xmark & \xmark & 62.5 & 0.1 & 27.9 & 1.3 & 26.6 & 1.8 & 57.4 & 2.6 & 41.1 & 15.6 & 11.8 \\
FRNet \cite{xu2023frnet} & 22.0 & 71.3 & \xmark & \xmark & \xmark & 0.6 & 0.1 & \xmark & \xmark & 65.2 & 0.2 & 47.4 & 1.9 & 52.3 & 2.0 & 76.8 & 2.2 & 59.5 & 14.5 & 24.3 \\
Cylinder3D \cite{zhu2021cylindrical} & 23.7 & 68.7 & \xmark & \xmark & \xmark & 0.8 & 1.7 & \xmark & \xmark & 64.0 & 0.3 & 44.3 & 1.1 & 57.5 & 8.5 & 78.8 & 16.4 & 57.9 & 24.6 & 25.3 \\
WaffleIron \cite{puy2023using} & 24.6 & 86.8 & \xmark & \xmark & \xmark & 0.1 & 1.5 & \xmark & \xmark & 60.8 & 0.9 & 43.6 & 0.5 & 55.1 & 4.8 & 82.0 & 18.7 & 67.6 & 25.5 & 19.2 \\ 
ours & \textbf{24.9} & 86.4 & \xmark & \xmark & \xmark & 0.2 & 1.2 & \xmark & \xmark & 61.9 & 0.5 & 47.3 & 0.5 & 55.9 & 4.7 & 80.4 & 18.5 & 66.9 & 24.9 & 23.4 \\
\hline
\end{tabular}
\end{center}
\end{table*}

\mycomment{
\begin{table}[t!]
\begin{center}
\caption{Semantic segmentation performance on SemanticKITTI validation set (sequence 8) of methods trained only on Sequence 04 (second column) or full training set (except sequence 8, third column).}
\begin{tabular}{ |c|c|c|}
\hline\
 method &   \textbf{Small data mIoU \%} & \textbf{Full data mIoU \%} \\ 
\hline
 Rangeformer \cite{kong2023rethinking} & 15.9 & 67.9 \\  
\hline
 FRNet \cite{xu2023frnet} & 22.5 & 68.7 \\ 
\hline
 Cylinder3D \cite{zhu2021cylindrical} & 23.2 & 64.3\\ 
\hline
 WaffleIron \cite{puy2023using} & 24.6 & 68.0 \\  
\hline
 ours & \textbf{24.9} & \textbf{69.0} \\
\hline
\end{tabular}
\end{center}
\label{tab:fine-tuning}
\end{table}
}

\subsection{Implementation Details}

We implemented our method in Python, C++, and CUDA. All the inference experiments were performed on a consumer laptop PC equipped with an AMD Ryzen 7 5800H CPU (3.2 GHz), 16GB of RAM, and Linux OS (the internal GPU has not been used). All the deep learning training were done on a Desktop PC with an Intel Core i9-10920X CPU (3.50~GHz), 32~GB of RAM, an Nvidia Titan RTX GPU, and a Linux OS. All the results reported refer to the performance obtained at the 45th epoch of the training process.

% \subsubsection{Inference time reduction}
\subsection{Inference Time Reduction: KDTree on GPU}
As outlined in Section \ref{sec:method}, the pre-processing module necessitates constructing a KDtree over the input point cloud. Initially implemented by the authors of WaffleIron using the KDTree class from the SciPy library \cite{2020SciPy-NMeth}, this approach proved fast on CPU but fell short of real-time requirements. Consequently, a GPU-boosted implementation emerged as a desirable alternative. By adapting cudaKDTree\footnote{\url{https://github.com/ingowald/cudaKDTree}}, an open-source library written in C++ and CUDA for constructing and querying KD-trees, we transitioned the computation to the GPU. Leveraging GPU optimization, the total time required for both building and querying the tree reduced drastically from approximately 160\,ms to a mere 15\,ms.

\subsection{Inference Time Reduction: Flattening Operation}
The original implementation of WaffleIron, in the spatial mix module, employs a batch matrix multiplication technique to average the features of points located within the same cell. This process involves constructing a large matrix in $\mathbb{R}^{B, HW, N}$, where $B$ represents the batch size, $HW$ denotes the flattened size of the grid under consideration, and $N$ is the total number of points of the point cloud. Subsequently, this matrix is multiplied by a weight matrix of size $\mathbb{R}^{N, B}$, with each element representing the inverse density of the corresponding cell that the point occupies.
\begin{table}[t!]
\begin{center}
\caption{Semantic segmentation performance on SemanticKITTI validation set (sequence 8) of methods trained only on Sequence 04 (second column) or full training set (except sequence 8, third column).}
\label{tab:val-big-data-skitti}
\begin{tabular}{ |c|c|c|}
\hline\
 method &  \textbf{Small data mIoU \%} & \textbf{Full data mIoU \%} \\ 
\hline
 Rangeformer \cite{kong2023rethinking} & 15.9 & 67.9 \\  
\hline
 FRNet \cite{xu2023frnet} & 22.5 & 68.7 \\ 
\hline
 Cylinder3D \cite{zhu2021cylindrical} & 23.2 & 64.3\\ 
\hline
 WaffleIron \cite{puy2023using} & 24.6 & 68.0 \\  
\hline
 ours & \textbf{24.9} & \textbf{69.0} \\
\hline
\end{tabular}
\end{center}
\end{table}
While technically accurate, this approach imposes significant demands on both time and memory resources since the initial matrix is highly sparse. The multiplication process, even when performed on a GPU, incurs substantial time overhead.
The required time of each of these multiplications considering all the $L=48$ layers is ~130\,ms.

By rearranging the sequence of these operations, we managed to reduce the total time required for all the layers to ~90\,ms, with a net advantage of ~40\,ms.
Instead of generating a large sparse matrix, we first perform an element-wise multiplication between the features of the point cloud and the weight matrix. Subsequently, we aggregate the weighted features using a scattering operation. This operation can be efficiently executed using the \textit{scatter\_add\_} function provided by the torch library \cite{pytorchlib}.
By supplying the indices of the cell to which each weighted point feature should be added, we rapidly construct the final matrix.
In the end, we were able to reduce the total runtime from ~365\,ms to ~180\,ms.
As described in Section~\ref{sec:ablation}, we obtained competitive results using only $L$=12 layers, but running \textit{real-time} in ~80\,ms.

\subsection{Datasets}
We evaluate our method on two extensive autonomous driving datasets: SemanticKITTI \cite{semantickittidataset} and nuScenes \cite{nuscenes}.

SemanticKITTI comprises 22 sequences, with each point cloud segmented into 19 semantic classes. We adopted the standard split where the initial 11 sequences formed the training set, with the exception of the 8th sequence used for validation. This sequence is the most complete and variegated. The remaining 11 sequences constituted the test set.

In nuScenes, each point is labeled with one of the 16 considered semantic classes. The dataset comprises 1000 scenes, but, following the official division, we used 700 scenes for training, 150 for validation, and 150 for testing. 

%\todo{If time, also real experiment with only qualitative results}

\subsection{Evaluation Metrics}
As commonly done in Semantic Segmentation tasks, we report the Intersection-over-Union (IoU) for class $i$ and the average score (mIoU) over all classes, where $\text{IoU}_i = \frac{\text{TP}_i}{\text{TP}_i+\text{FP}_i+\text{FN}_i}$
. $\text{TP}_i$
, $\text{FP}_i$ and $\text{FN}_i$ are the true positives, false positives, and false negatives.

\subsection{Performance on Autonomous Driving Datasets}

\begin{table*}
\begin{center}
\caption{Semantic segmentation performance on nuScenes validation set. \textsuperscript{*}: Regarding 2DPASS, we report the results of the
baseline of \cite{yan20222dpass} trained with lidar data but no images, i.e., in the same setting as the other methods in this table.}
\label{tab:test-nuscenes}
\begin{tabular}{c p{2.5mm} | p{2.5mm}p{2.5mm}p{2.5mm}p{2.5mm}p{2.5mm}p{2.5mm}p{2.5mm}p{2.5mm}p{2.5mm}p{2.5mm}p{2.5mm}p{2.5mm}p{2.5mm}p{2.5mm}p{2.5mm}c}
\hline 
\rb{method} & \multicolumn{1}{c}{\rb{\textbf{mIoU \%} }} &
\rb{barrier} & \rb{bicycle} & \rb{bus} & \rb{car} & \rb{const. veh.} & 
\rb{motorcycle} & \rb{pedestrian} & \rb{traffic cone} & \rb{trailer} & \rb{truck} & \rb{driv. surf.} & \rb{other flat} & \rb{sidewalk} &
\rb{terrain} & \rb{manmade}  & \rb{vegetation} \\ 
\hline
RangeNet++ \cite{milioto2019rangenet++} & 65.5 & 66.0 & 21.3 & 77.2 & 80.9 & 30.2 & 66.8 & 69.6 & 52.1 & 54.2 & 72.3 & 94.1 & 66.6 & 63.5 & 70.1 & 83.1 & 79.8 \\
PolarNet \cite{zhang2020polarnet} & 71.0 & 74.7 & 28.2 & 85.3 & 90.9 & 35.1 & 77.5 & 71.3 & 58.8 & 57.4 & 76.1 & 96.5 & 71.1 & 74.7 & 74.0 & 87.3 & 85.7 \\
SalsaNext \cite{cortinhal2020salsanext} & 72.2 & 74.8 & 34.1 & 85.9 & 88.4 & 42.2 & 72.4 & 72.2 & 63.1 & 61.3 & 76.5 & 96.0 & 70.8 & 71.2 & 71.5 & 86.7 & 84.4 \\
Cylinder3D \cite{zhu2021cylindrical} & 76.1 & 76.4 & 40.3 & 91.2 & 93.8 & 51.3 & 78.0 & 78.9 & 64.9 & 62.1 & 84.4 & 96.8 & 71.6 & 76.4 & 75.4 & 90.5 & 87.4 \\
2DPASS\textsuperscript{*} \cite{yan20222dpass} & 76.2 & 75.3 & 43.5 & 95.3 & 91.2 & 54.5 & 78.9 & 72.8 & 62.1 & 70.0 & 83.2 & 96.3 & 73.2 & 74.2 & 74.9 & 88.1 & 85.9 \\
WaffleIron \cite{puy2023using} & 77.6 & 78.7 & 51.3 & 93.6 & 88.2 & 47.2 & 86.5 & 81.7 & 68.9 & 69.3 & 83.1 & 96.9 & 74.3 & 75.6 & 74.2 & 87.2 & 85.2 \\
ours & \textbf{77.6} & 78.5 & 49.6 & 91.8 & 87.6 & 52.7 & 86.7 & 82.2 & 70.1 & 67.2 & 79.7 & 97.0 & 74.7 & 76.8 & 74.9 & 87.5 & 85.0 \\
\hline
\end{tabular}
\end{center}
\end{table*}

\begin{table}[t!]
\begin{center}
\caption{Small data setup. Results on methods trained ONLY on sequence 04 of SemanticKITTI (271 point clouds) and tested on sequence 08 (4071 point clouds). }
\label{tab:small-data}
\begin{tabular}{ |c|c|c|c|c|c|c|}
\hline\
 method & range & L & drop & skip & aug & \textbf{mIoU \%} \\ 
\hline
 original & \xmark & 48 & - & - & \xmark & 24.4 \\   
 ours & \cmark & 48 & \xmark & \xmark & \xmark & \textbf{24.9} \\
 % ours & \cmark & 48 & \cmark & \xmark & \xmark & - \\
 ours & \cmark & 48 & \cmark & \cmark & \xmark & 22.3 \\
 ours & \cmark & 48 & \xmark & \cmark & \xmark & 23.8 \\
 \hline
 original & \xmark & 48 & - & - & \cmark & 29.3 \\ 
 ours & \cmark & 48 & \xmark & \xmark & \cmark & 32.7 \\ % fun fact. Non fa differenza o.O
 ours & \cmark & 48 & \xmark & \cmark & \cmark & \textbf{32.7} \\
 % ours & \cmark & 48 & \xmark & \cmark * & \cmark & \textbf{33.0} \\
 \hline
 original & \xmark & 12 & - & - & \cmark & 28.4 \\  
 ours & \cmark & 12 & \cmark & \cmark & \cmark & 20.6 \\
 ours & \cmark & 12 & \xmark & \xmark & \cmark & 28.5 \\
 ours & \cmark & 12 & \xmark & \cmark & \cmark & \textbf{29.8} \\
\hline
\end{tabular}
\end{center}
\end{table}

\begin{table}[t!]
\begin{center}
\caption{Results on methods trained on full training set of SemanticKITTI and tested on sequence 08 (validation set). Network configuration with only 12 layers and active dataset augmentation. }
\label{tab:12l}
\begin{tabular}{ |c|c|c|c|c|c|c|}
\hline\
 method & range & L & drop & skip & aug & \textbf{mIoU \%} \\ 
\hline
 original & \xmark & 12 & - & - & \cmark & 62.6 \\  
 \hline
 ours & \cmark & 12 & \xmark & \cmark & \cmark & \textbf{65.8} \\
\hline
\end{tabular}
\end{center}
\end{table}

The performance evaluation is conducted in two different setups.
The first setup aims to evaluate the generalization capabilities of the models in scenarios where data is scarce. Referred to as the ``small data'' setup, this approach recognizes that while a vast amount of available data is typically beneficial for deep learning models, it may not be sufficient to truly evaluate their efficiency. Even a well-trained model on a large dataset may not have an optimal architecture, yet still deliver excellent results on the test set. To address this, we initially train using only a single, concise set of point cloud data. Specifically, these trainings are conducted solely on SemanticKITTI sequence 04, with performance validation on sequence 08. Sequence 04 is the smallest of the training dataset, consisting of only 271 point clouds. In contrast, sequence 08 comprises 4071 point clouds, making the models' learning very challenging. Data augmentations applied in this setup include random X-Y flipping and/or rotation, along with random scaling with a maximum scale factor of 0.1.

%We run these experiments to fine-tune the parameters for the full dataset trainings. This setup is called ”Full Dataset”.
We retrained the models to validate their performance on the "full data" setup, where the full datasets are utilized.
The trainings were conducted on both SemanticKITTI and nuScenes datasets, with the latter exclusively utilized in the full data setup. Our models were trained with $L$ = 48 layers.
We employed 256-dimensional point features and a grid resolution of ~40\,cm for SemanticKITTI, while for nuScenes, we used $F$ = 384 and a grid resolution of ~60\,cm.
Additionally, exclusively for SemanticKITTI, following the data augmentation applied in the small data setup, we incorporated the InstanceCutMix and PolarMix techniques utilized by the authors of WaffleIron.

The results of the small data setup are reported in Table~\ref{tab:val-little-data-skitti}. It shows a clear improvement of our method with respect to all the other state-of-the-art methods, including the original architecture of WaffleIron. This is due to the much more informative view given from the range image, which helps the model to better share the features during the Waffle processing. It is able to improve the results of 0.3\% of mIoU with respect to the original WaffleIron model.

Our method is also the best performing in the validation sequence of the full data setup (see Table~\ref{tab:val-big-data-skitti}), in which it obtains a mIoU of 69.0\% \new{and improves the results of the original WaffleIron of 1.0\%}. This confirms the generalization capabilities that the model obtains in the small data setup. In Section~\ref{sec:ablation}, we conduct a more in-depth analysis of the various components that led us to the architecture development.

%In Table~\ref{tab:test-skitti}, we report the results on the \textit{test} set of SemanticKITTI, where all models are trained on the full Training set, including Sequence 08. It can be seen that the original WaffleIron and RangeFormer obtain higher results with respect to our models, but this is still competitive and better than many state-of-the-art models such as 2DPASS (trained only on LiDAR point clouds, without making use of the RGB images), CENet and so on.

%In Table~\ref{tab:test-nuscenes}, we report the results on the \textit{val.} set of nuScenes. The table clearly indicate the superiorness of both the original WaffleIron and our method, which obtains slightly better performances on average. The benefit is not as large as in SemanticKITTI probably because the range image is here applied to point clouds obtained from a LiDAR having 32 channels instead of 64 channels as in SemanticKITTI. The range image hence is less informative and the objects are less distinguishable. 
%It's interesting to note that fundamental semantic objects like motorcycle, pedestrian and traffic cones are better segmented by our method. Also, we reach the top results in construction vehicle, drivable surface, sidewalk, terrain and manmade classes.
In Table~\ref{tab:test-nuscenes}, we present the results obtained on the validation set of nuScenes. The table clearly indicates the superiority of both the original WaffleIron and our method, with our approach achieving slightly better performance on average. However, the observed benefit is not as pronounced as in SemanticKITTI. This discrepancy may be attributed to the fact that the range image projection is applied to point clouds obtained from a LiDAR with 32 channels, compared to the 64 channels used in SemanticKITTI. Consequently, the range image provides less informative data, making object distinctions less clear. 
It's worth noting that our method demonstrates improved segmentation of fundamental semantic objects such as motorcycles, pedestrians, and traffic cones. Furthermore, we achieve top results in classes related to construction vehicles, drivable surfaces, sidewalks, terrain, and manmade structures.

\section{ABLATION STUDY}\label{sec:ablation}

%In this section, we are going to describe the steps that let us design and identify the most suitable architecture to train on the \new{"big data"} setup, which takes time and consume a lot of energy. These experiments are conducted only on SemanticKITTI dataset.
In this section, an ablation study conducted on the SemanticKITTI dataset is carried out and described.

Table~\ref{tab:small-data} reports the results of the experiments on the small data setup. On the top block, we only applied data augmentations such as random X-Y flip and/or rotation and random scaling, with a maximum scaling factor of 0.1.
The columns drop, skip and aug refer, respectively, to three techniques that we tested. The first is a random dropout of the neighbor points used in the embedding module.
The second is the application of a skip connection between the neighbor features computed from the embedding module and the processed points computed by the backbone. In other words, we just sum up the neighbors embedding and the final point cloud processing just before the final classification of the points.
The third is the application of the data augmentation techniques InstanceCutMix and PolarMix that WaffleIron authors used. It's important to note that these introduce also instances that are not present in the sequence 04 that we use for training, taking these instances from the other sequences. These tests are a good indicator of the performance of the methods on the full training set.

From the results shown in Table~\ref{tab:small-data}, it can be observed that the dropout technique does not significantly improve the robustness of the model. However, skip connections exhibit interesting behavior: while they marginally decrease performance without augmentation, they lead to performance enhancement when InstanceCutMix and PolarMix augmentations are applied, yielding a mIoU of 32.7\%.

Furthermore, we analyzed the performance of the method when using a reduced number of layers ($L=12$). The results, summarized in Table~\ref{tab:12l}, mirror the observations made with the full model configuration: dropout tends to decrease performance, whereas skip connections tend to enhance it.
Motivated by these findings, we conducted a comprehensive study with $L=12$ layers on the entire training set, validating the results on sequence 08. Remarkably, our method achieved a mIoU of 65.8\%, surpassing the baseline method WaffleIron's mIoU \new{(62.6\%)}. This performance is comparable to methods employing $L=48$ layers. Additionally, our method meets real-time requirements with a runtime of less than ~100\,ms, averaging at ~80\,ms.

\section{CONCLUSIONS}

%We have addressed the challenge of achieving real-time semantic segmentation of point clouds \new{in a setup in which the available dataset is small} by combining the strengths of two state-of-the-art existing models.
We addressed the challenge of achieving real-time, accurate semantic segmentation of point clouds in cases where the training dataset is small.
Building upon two existing state-of-the-art models, our architecture proficiently leverages local feature extraction during point cloud embedding and hybrid projection of both the 3D point cloud and the range image, thus increasing the accuracy and efficiency.
%Our architecture proficiently exploits local feature extraction during point cloud embedding and the range image introduction, optimizing both accuracy and efficiency. 
Through extensive analysis on benchmarks like SemanticKITTI and nuScenes, our experiments show that incorporating local features into the embedding module and integrating range image data greatly boosts performance in situations with scarce data. Additionally, our method also improves results in data-rich contexts, exhibiting its versatility across various data scenarios. Through code optimization we significantly decreased the system runtime, providing a robust solution for real-world semantic segmentation tasks for autonomous robotics. 
% significant runtime reductions without sacrificing performance.
% Our open-source implementation facilitates further research and application in fields like autonomous driving and robotics. Overall, our work provides a robust solution for real-world semantic segmentation tasks.

% \addtolength{\textheight}{-12cm}   % This command serves to balance the column lengths
%                                   % on the last page of the document manually. It shortens
%                                   % the textheight of the last page by a suitable amount.
%                                   % This command does not take effect until the next page
%                                   % so it should come on the page before the last. Make
%                                   % sure that you do not shorten the textheight too much.

%%%%%%%%%%%%%%%%%%%%%%%%%%%%%%%%%%%%%%%%%%%%%%%%%%%%%%%%%%%%%%%%%%%%%%%%%%%%%%%%

%%%%%%%%%%%%%%%%%%%%%%%%%%%%%%%%%%%%%%%%%%%%%%%%%%%%%%%%%%%%%%%%%%%%%%%%%%%%%%%%

%%%%%%%%%%%%%%%%%%%%%%%%%%%%%%%%%%%%%%%%%%%%%%%%%%%%%%%%%%%%%%%%%%%%%%%%%%%%%%%%
% \section*{APPENDIX}

% Appendixes should appear before the acknowledgment.

% \section*{ACKNOWLEDGMENT}

% \todo{The preferred spelling of the word acknowledgmentÓ in America is without an eÓ after the gÓ. Avoid the stilted expression, One of us (R. B. G.) thanks . . .Ó  Instead, try R. B. G. thanksÓ. Put sponsor acknowledgments in the unnumbered footnote on the first page.}

%%%%%%%%%%%%%%%%%%%%%%%%%%%%%%%%%%%%%%%%%%%%%%%%%%%%%%%%%%%%%%%%%%%%%%%%%%%%%%%%

\bibliographystyle{IEEEtran}
\balance
\bibliography{bib}

\end{document}